# Answer Identification in Collaborative Organizational Group Chat

Naama Tepper, Naama Zwerdling, David Naori* Inbal Ronen
IBM Research – Haifa, Israel   *Technion, Israel
*{naama.tpper,naamaz,inbal}@il.ibm.com* *dnaori@cs.technion.ac.il

**Abstract.** We present a simple unsupervised approach for answer identification in organizational group chat. In recent years, organizational group chat is on the rise enabling asynchronous text-based collaboration between co-workers in different locations and time zones. Finding answers to questions is often critical for work efficiency. However, group chat is characterized by intertwined conversations and 'always on' availability, making it hard for users to pinpoint answers to questions they care about in real-time or search for answers in retrospective. In addition, structural and lexical characteristics differ between chat groups, making it hard to find a 'one model fits all' approach. Our Kernel Density Estimation (KDE) based clustering approach termed *Ans-Chat* implicitly learns discussion patterns as a means for answer identification, thus eliminating the need for channel-specific tagging. Empirical evaluation shows that this solution outperforms other approaches.

## Introduction

Modern group work is characterized by co-workers collaborating on shared tasks across different geographic locations and different time zones. Group chat tools, such as Slack and Microsoft Teams, have become essential for enabling asynchronous instant communication in organizations. However, their 'always on' nature comes with a major cost of information overload, lack of ability to track relevant content and specifically find answers to questions. This problem even worsens when trying to catch-up on content after being away for a while. Many of the conversation patterns in organizational group chat include posting of questions, such as where to find information, whether a server is up, who is an expert for a problem, and more. Finding answers to those question either in real-time or when searching for them at a later point can be crucial for efficient task performance in the workplace. Thus, automatically identifying answers to questions in a chat can be very crucial. It can enable among others alerting on the arrival of an answer to a posted question (when someone posts a potential answer), the addition of these answers to conversation summaries to assist in catching up after some absence, and the creation of Q&A databases for conversational bot systems for future look-ups and recommendations of answers. Notably, Slack introduced the ability to respond in threads, similar to discussion board topics and posts. By adding more structure, these should have made the identification of answers easier for users, however, threads are not widely used. They create overhead for users having to open each thread separately without a clear overview of what has been going on in them. After leaving a threaded reply, a user must manually find the thread and expand it. In addition, as threads are hidden, they also discourage participation. Furthermore, users can find it hard to decide which thread their answer should be posted to, as sometimes it might fit to more than one thread.

Group chat lingo, communication forms and content differ from other forms of textual communication such as discussion boards. Particularly, group chat is poor at managing interruptions and turn taking with interleaved conversations on different discussion topics (Uthus and Aha, 2013). Thus, group chat is poor at conveying comprehension between users. The mention ('@' sign) feature aids in chat orientation by allowing users to specifically refer to someone in the group chat. Still, users can find it difficult to focus their attention and notice the answers to questions of importance to them, especially if they have been away for more than just a short period of time. Additional differences between group chats and discussion forums are provided in Table I. These differences, especially the unique characteristics of chat text, present numerous challenges for text analysis and in particular answer identification tasks (O'Neill and Martin, 2003; Uthus and Aha, 2011; Hu et al., 2012; Lewis and D., 1992; Berkovsky et al., 2008; Tepper et al., 2018a; Tepper et al., 2018b). Specifically, interleaved conversations make it hard to understand which messages belong to the same thread of conversation and thus to identify which question a message is an answer of. Table II depicts an example of interleaved chat conversations with questions and their answers: Yosef is asking a question which is answered by Jessi. However, their discussion is interrupted by a conversation between Jessi and Jon.

Most of the related previous work on answer identification focused on discussion boards or Q&A forums and not on group chat. Furthermore, most of the approaches used are supervised methods requiring labeled training corpora (Wang et al., 2011; Zhang et al., 2017; Isiaka Obasa et al., 2016; Catherine et al., 2013; Kim et al., 2010; Hong and Davison, 2009). These methods demonstrate that structural features are best predictors for discourse act. Specifically, a paper by Zhang et al. (2017) studied the task of classifying comments of Reddit discussions into a set of discourse acts (including question, answer, announcement, agreement, and more). They compared unstructured prediction models (Logistic Regression) and structured prediction models considering the sequence of comments (Hidden Markov Model – HMM and Conditional Random Field - CRF). Both structured prediction models performed better than the logistic regression model, demonstrating the importance of context for prediction and achieving a F1 score of 0.75.

Table I. Group chat vs. discussion forums

|  | **Group chat** | **Discussion forums** |
|---|---|---|
| **Conversation structure** | Intertwined discourses | Segregated discourses |

| Language | Poorly formed | Well formed |
|---|---|---|
| **Question position** | Unknown | First post |
| **Participants** | Colleagues | Mostly strangers and 'power users' |
| **Context** | Partial | Self-contained |
| **Discussion content** | Few answers | Mostly answers |

Unsupervised approaches for answer identification showed mixed results (Perumal and Hirst, 2016; Perumal, 2016; Cong et al., 2008; Deepak and Visweswariah, 2014). Perumal (2016) claimed that "*Experimental results indicate that purely unsupervised methods are not adequate for tackling a task as complex as forum post categorization*." On the other hand, Deepak & Visweswariah (2014) introduced a method relying only on textual features and applying an iterative approach modeled on the EM meta-algorithm (Dempster, Laird, & Rubin, 1977). Their method achieved a reasonable f-score of 0.64 when applied on discussion forums. However, since chat text differs in its characteristics from discussion posts, the prospects of using only textual analysis is questionable.

Given questions in an organizational group chat channel feed, we want to find the set of messages in the feed which answer these questions. For this purpose, we present *Ans-Chat*, a clustering approach for answer identification in goup chat. *Ans-Chat* utilizes Kernel Density Estimation (KDE) in addition to a conservative approach for initialization and update. Experimental validation using our unsupervised approach achieved an F-score of 0.714 on average over two chat channels.

Table II. Group chat example of interleaved questions (yellow) and their answers (green). Parentheses identify the row of the question the answer refers to.

| # | Message Type | User | Text |
|---|---|---|---|
| 1 | | Jessi | <@Jon>: The fixed 10.0.0.3 build is ready: <https://www.example.com/> 8.2-5 should be next. then 9.0.0.7 11.0.1.0 was hit hardest by issues with the build slaves after that power outage |
| 2 | Question | Yosef | Are you working from the conference? |

| 3 | | Jon | <@Jessi>: ok Thanks |
| 4 | Answer (2) | Jessi | I am no longer working from the conference because I can't get the VPN to connect reliably I'm at starbucks I'll head back once these builds are done or they kick me out |
| 5 | | Yosef | lol |
| 6 | | Jessi | yeah. It turns out when you tell 200+ software developers the hotel wifi password if falls over |
| 7 | | Yosef | I noticed that |
| 8 | Question | Jon | <@Jessi>: what test case do you suggest to test first? Install old 10.0.0.3 release build < configure> update to build 895 2. Install old 10.0.0.3 release build < configure> update to 11.0.2.0 |
| 9 | Answer (8) | Jessi | <@Jon>: Install new 10.0.0.3 build and upgrade to 11.0.2.0 |
| 10 | | Jon | ok |
| 11 | Answer (8) | Jessi | installing the old 10.0.0.3 build (or old 9.0.0.7 build (or 8.2-5 build &.))) and upgrading won't work until I finish the mitigation script because it's the old installers that have the issues |
| 12 | | Jon | right gotcha |

# Algorithm

In this section we present *Ans-Chat*, an unsupervised algorithm which given a set of identified questions in a feed of messages of an organizational group chat, pairs each question with its potential answers out of subsequent messages within a given window *w* of messages. We start by pairing each question message with each of its subsequent messages. Each subsequent message is viewed as a *potential answer*. Next, we cluster each of the message pairs (question and potential answer message) into one of two clusters – the 'answers' cluster *A* or to the 'non-answers' cluster *N*.

Different clustering methods differ in (i) choice of objective function; (ii) probabilistic generative model used to describe the data; and (iii) heuristics used. In this paper we show our modifications and enhancements over the classic k-means and the algorithm presented by Deepak & Visweswariah (2014) relying on Kernel Density Estimation (KDE) for cluster estimation and representation and utilizing group-chat specific domain knowledge to enhance the clustering heuristic using a conservative mechanism for initialization and update of clusters.

In the following we call the person posing a question the *asker* and the person who posts an answer the *answerer*.

## General Clustering Algorithm

We pair each question with subsequent messages each of which we view as a potential *answer*. The algorithm assigns each pair *x = (question, potential answer)* to one of two clusters: the *answers* cluster *A* and the *non-answers* cluster *N*, such that $X = A \cup N$ are all *(question, potential answer)* pairs in the channel. Note that a question can have multiple answers in *A*.

Given cluster assignments of message pairs for *A* and *N*, models $M_A$ and $M_N$ can be learned. These models function as scoring mechanisms by which later cluster assignment is made. For example, in the k-means algorithm (Jain, 2010) the mean of all datapoints is calculated per cluster, and $M(x)$ outputs the Euclidian distance of *x* from that mean value.
A high-level description of the basic clustering algorithm is given by:

Initialization:
- Initialize *A* with a heuristic *H*, $A = \{x \in X | H(x) = True\}$ where *H*(x) is a simple heuristic function. *N=X/A*.
- Learn a models $M_A$ and $M_N$ for *A* and *N* such that $M_A(x)$ and $M_N(x)$ evaluate the conformance of pair *x* with the corresponding model.

Iterate for n iterations or until convergence:
- Re-classify the pairs in *X* according to the current models:
  $A = \{x \in X | M_A(x) > M_N(x)\}, N = X/A$
- Re-learn the models $M_A(x)$ and $M_N(x)$ according to the (possibly) new sets.

## Kernel Density Estimation (KDE) Model

In this section we describe a probabilistic model *M* used for scoring the conformance of a message pair to a cluster. We used the non-parametric Kernel Density Estimation (KDE) (Silverman, 2018). Specifically, given cluster assignments for *A* and *N*, we used KDE to consolidate the features and estimate probability density functions $f_A$ and $f_N$ serving as $M_A$ and $M_N$ respectively thus providing a measure of conformance (score) of a message pair and a cluster.

The KDE method can be described simplistically as a smooth histogram – the most basic non-parametric method to represent data. KDE takes this representation a step further, not relying on bin size and representing each datapoint as a smooth function (usually standard normal). Thus, making the entire data representation smooth. In addition, a smoothing bandwidth $\sigma$ is used as input or learned to best represent the data. Notably, this approach allows modeling when the form of the density is not known, and all examples are used in the estimation.

For each cluster in each iteration, we fit a KDE to obtain a generative model of that cluster. This allowed to compute a likelihood $P_y(x)$ for any question-message pair *x* and cluster *Y*, evaluating the conformance of pair *x* with the corresponding cluster *Y*. $P_y(x)$ is given by:

$$P_y(x) = \frac{1}{y} \sum_{y \in Y} K\left(\left\|\frac{x-y}{\sigma}\right\|^2\right)$$

where $Y$ is the cluster of datapoints, $K$ is the kernel used and $\sigma$ is the smoothing bandwidth.

Our approach differs from previous approaches (Deepak & Visweswariah, 2014) firstly by applying non-parametric KDE to evaluate cluster distribution per each iteration of the clustering algorithm, and secondly, by using KDE probability function as a scoring function by which cluster assignment is made; thirdly, differently from other clustering approaches, here, cluster representation includes all members of the cluster rather than a single representation such as mean, median or mode (Chaturvedi, Foods, Green, & Carroll, 2001).

The KDE parameterization used is detailed in the 'Experimental Setup' section.

## Conservative Clustering

Convergence is a known pain-point in clustering algorithms: we would like cluster assignment robustness to increase with the number of clustering iterations. However, in many cases, and specifically when the number of features is large, datapoints tend to bounce between different clusters not converging into a specific cluster. To address this issue, we defined a new heuristic for initialization and clustering tailored to domain specific settings (in our case, the group chat domain) in order to ensure convergence of our algorithm and improve performance. We term the heuristic 'conservative clustering'. Two small 'seed' clusters $A', N' \subset X$, were defined containing only pairs with high level of confidence, which do not move throughout the algorithm execution. Notably, in this setting only part of the message pairs are initially clustered into the $A$ and $N$ clusters: $A'$ datapoints are constantly clustered into cluster $A$, and $N'$ datapoints are constantly clustered into cluster $N$, while the remaining messages are left un-clustered in the initialization step and are re-assigned in every iteration of the algorithm. For example, a message pair $a \in A'$ is bound to be only in cluster $A$, cannot move to cluster $N$ and is always used (together with the other data points assigned to $A$) to estimate $M_A$. This is in contrast to a basic clustering approach which: (i) initializes $A$ and $N$ with all channel's pairs, and (ii) all pairs are free to move between clusters. To construct $A$' and $N$' we used group chat domain knowledge: We constrained the initial seed cluster for answers $A$' to message pairs where the asker mentioned the answerer or the asker later acknowledged the answerer (such as "ok" or "yes"), as these have high probability of being answers to the question. $N$' was initialized with pairs where question and potential answer originated from the same author or where time between both messages was improbable for being an answer to the question (see Experimental Setup section).

## Ans-Chat Algorithm

We refer to the combination of the conservative approach and the KDE modeling as *Ans-Chat*.

The following describes our *Ans-Chat* algorithm in high level:

Initialization:
- Initialize A' and N' seed clusters as described above
- Initiate probability density functions $f_A$ and $f_N$ and their bandwidth σ using KDE on A' and N' respectively.

Iterate for n iterations or until convergence of A and N:
- Re-classify the pairs in $X/(A' \cup N')$ according to the current models: = $\{x \in X | P_A(x) \geq P_N(x)\}, N = X/A$
- Re-learn probability density functions $f_A$ and $f_N$ using KDE according to the (possibly) new sets.

## Structural and Textual Feature Set

For each pair of a question and its potential answer we extracted textual and structural features. As mentioned earlier, chat text is sparse, and applying standard text analysis for the extraction of textual features usually achieves poor results. Therefore, we focused on a simple textual Jaccard similarity score to capture the degree of overlap between a question and a potential answer. We used the messages' stemmed tokens after removing stop-words. For structural features, we used several measures to capture the distance between the messages and between their authors. We chose the following structural features: (1) The distance in number of messages between the question and the potential answer. (2) The distance in number of messages between the potential answer and the questioner's next message. (3) Whether the questioner mentions the replier in the question message. Additional features were used to construct Singular Value Decomposition (SVD) (Golub & Reinsch, 1970) representations for dimensionality reduction. The full list of features is provided in Table IIII. Note that each row represents a feature category which might include multiple features as described in the Features column.

Table III. Full feature set

| Category | Features |
|---|---|
| **Structural features** | |
| Distance of the potential answer from the question | Distance in number of messages or time difference in seconds between the question and the potential answer |
| Mentions | Whether the asker mentioned the answerer in the question or in another message in the same window; whether the answerer mentioned the asker in the potential answer, or the asker mentioned a different user than the answerer in the answer |
| Asker's response after the potential answer | Whether the potential answer is followed by a message of the asker, particularly if it contains gratitude or acknowledgment ("okay", "thanks", "thumbs up", etc.) |
| Messages position | Position of last message of the asker/answerer in the window and the potential answer's relative position in the window; distance in messages or time difference from the previous/next answerer/asker message |
| Answerer activity in the window | Proportion of messages by the answerer in the window, normalized or not by the average answerer activity in the entire feed or normalized by the window activity |
| Previous correspondence and activity | Number of messages in which the asker and answerer mentioned each other in the feed or authored a message in previous windows |
| **Textual Features** | |
| Textual similarity between question and potential answer | The cosine or Jaccard similarity using stemmed tokens (removing stop words); Each of the above features was implemented as a raw feature or normalized by the window average |
| Textual similarity between question and potential answer using word Embedding | We used Stanford's GloVe word embedding (Pennington, Socher, & Manning, 2014) trained over Google news (Aboufadel et al., 2013) and Twitter (Godin, Vandersmissen, De Neve, & Van de Walle, 2015) collections. Word embedding vectors of the question messages were summed and summed embedding vectors of the potential answer messages were then subtracted from that sum. Each of the above features was implemented as a raw feature or normalized by the window average |
| Text length | Length of question, potential answer and their ratio |
| Text type | Whether the text of the potential answer contained identified categories as identified by eAssistant (Nezhad et al., 2017). The identified categories are: promise, request and actionable statements. |

# Experimental Evaluation

## Tagged Dataset

We evaluated our answer identification approach on two Slack channels of work groups from our organization, denoted as channel 1 and channel 2, respectively. Data was collected over a period of one year, with over 20,000 messages altogether. Channel 1 was characterized by a relatively high number of questions – 13% of the messages compared to only 5% of question messages in channel 2. However, channel 2 showed higher user engagement with a median of 30 messages per user compared to only 13 messages per user in channel 1. The channels' statistics are summarized in Table IV. Questions were identified using a combination of the eAssistant web service (Nezhad et al., 2017) and Watson Natural Language Classifier[1].

Table IV. Tagged dataset statistics

|  | channel 1 | channel 2 |
| --- | --- | --- |
| **#messages** | 8,345 | 15,052 |
| **#users** | 85 | 62 |
| **Median #messages per user** | 13 | 30 |
| **#questions** | 1,105 (13%) | 2,854 (5%) |

For each channel, 1000 messages were tagged, from which about 300 identified questions and their answers were tagged per channel. Each message was tagged by three different taggers. Taggers indicated for each question which of the messages following it contained its answers. The agreement between the taggers (Fleiss kappa) was 0.741 and 0.758 on channels 1 and 2 respectively. All comparisons in the following sections refer to the majority of the 3 taggers.

## Experimental Setup

For construction of question-potential answer pairs, we chose a window size $w=10$ of messages following each question. We ran the algorithm with maximal iteration number of $n=15$ or until less than 10 pairs switched clusters. Improbable time for an answer was defined as less than a second or more than 10 hours. The algorithm was implemented in python using sklearn for KDE and k-means, nltk for text processing and language and translation models, and gensim for word2vec. For KDE, we used 10 bandwidth points and 3-fold cross validation to choose the bandwidth for the smoothing parameter.

---

[1] https://www.ibm.com/watson/services/natural-language-classifier/

Results

As previous methods used solely textual features to find answers to questions in group chat, we first compared Deepak and Visweswariah's (2014) model to our KDE modeling approach (without conservative clustering) over only textual features. We then evaluate the performance of the *Ans-Chat* algorithm with and without the conservative clustering variant. Finally, we evaluated the contribution of structural features vs. textual features and tested a structural-textual hybrid approach.

**Choosing Modeling Approach**

As stated above, we first compared Deepak and Visweswariah's (2014) model to our KDE model using only a single textual feature (and without conservative clustering). We implemented state-of-the-art approaches as our baseline using their language, translation and the combined model (lambda=0.5) for which they received best results. For the KDE-based model we used Jaccard distance between the question and the potential answer. Jaccard distance was computed based on the message tokens as described in the 'Structural and Textual Feature Set' section. Table V shows the results for different settings of previous state-of-the-art and our KDE model. Channel 1 got the highest F-score results in all setting compared to channel 2. Our KDE based clustering dominates the precision and F-score results in both channels achieving F-scores of 0.314 and 0.301 on channel 1 and channel 2 respectively.

Table V. Quality evaluation of different configurations from Deepak and Visweswariah (2014) and our KDE approach

|  | channel 1 | | | channel 2 | | |
|---|---|---|---|---|---|---|
|  | Precision | Recall | **F-score** | Precision | Recall | **F-score** |
| Translation | 0.125 | **0.996** | 0.222 | 0.112 | **0.995** | 0.202 |
| Language | 0.125 | 0.836 | 0.218 | 0.121 | 0.735 | 0.207 |
| Lang.+Trans.0.5 | 0.124 | 0.871 | 0.218 | 0.119 | 0.850 | 0.209 |
| KDE with Text (Jaccard) | **0.204** | 0.681 | **0.314** | **0.202** | 0.585 | **0.301** |

**Determining Clustering Variant**

Establishing that KDE with Jaccard is a better than the previous textual approach, we now tested the contribution of conservative clustering to overall performance. Table VI details the differences between standard clustering and our conservative clustering variant over the Jaccard textual feature. As we can see in both channels the conservative clustering approach dominates the regular clustering approach in almost all quality measures (recall, precision and F-score).

Table VI. Quality evaluation of KDE with Text (Jaccard) with and without conservative clustering

|  | channel 1 | | | channel 2 | | |
|---|---|---|---|---|---|---|
|  | Precision | Recall | **F-score** | Precision | Recall | **F-score** |
| Regular | 0.204 | 0.681 | 0.314 | **0.202** | 0.585 | 0.301 |
| Conservative | **0.273** | **0.741** | **0.399** | 0.194 | **0.820** | **0.314** |

**Choosing the Feature Set**

Organizational group chat is also characterized by structural features other than the text of the messages. For example, users can also mention other users in their messages. The mentioned users will get a special notification when someone else mentions them. In this section we try to utilize chosen structural features described in the 'Structural and Textual Feature Set' section as well as textual features to improve the (poor) textual-based results. Table VII describes the clustering results using the complete *Ans-Chat* settings - conservative clustering and KDE modeling - over different sets of features. Table VII demonstrates that adding these chosen structural features data improved the results significantly. Channel 2 even doubled its F-score.

Notable, we also examined a larger feature set. As the number of features was large, representation of each message pair became computationally expensive. These features were used to construct Singular Value Decomposition (SVD) (Golub & Reinsch, 1970) representations for dimensionality reduction as described in the 'Structural and Textual Feature Set' section. However, the addition of these features reduce performance achieving F-score of 0.585 and 0.578.

We also tested if combining these chosen textual and structural features would yield better results and indeed the textual and structural features was the highest in both channels.

Table VII. Quality evaluation over different features with conservative clustering

|  | channel 1 | | | channel 2 | | |
|---|---|---|---|---|---|---|
|  | Precision | Recall | **F-score** | Precision | Recall | **F-score** |
| *AnsChat* with Text (Jaccard) | 0.273 | **0.741** | 0.399 | 0.194 | **0.820** | 0.314 |
| *AnsChat* with Structure | **0.787** | 0.509 | 0.618 | **0.735** | 0.595 | 0.657 |
| *AnsChat* **with Text + structure** | 0.737 | 0.711 | **0.724** | 0.654 | 0.765 | **0.705** |

# Discussion and Conclusions

In this work we described a method for answer identification in an organizational collaborative environment of group chat. By implicitly modeling communication characteristics per each different group chat feed, our approach demonstrated improved performance over state-of-the-art. Specifically, our analysis and results indicated that standard textual features provided limited value to the analysis of group chats due to chat text unique characteristics. However, structural features were shown to have great impact and improved results in all settings.

Detecting question chat messages is a non-trivial task (Isiaka Obasa, Salim, & Khan, 2016). The technologies we used identified question in group chat. However, they also presented false positive and false negative results – some questions were not tagged as questions while some general messages were tagged as questions. Some examples are shown in Table VIII. These inconsistencies might have negatively affected our algorithm performance.

Table VIII. False positive and false negative results in question identification.

| **False positive** | **False negative** |
| --- | --- |
| Yup can ping it | Should I still be working on the end-to-end deployment stuff? |
| Ah, ok- then I will | Is it running now? |
| I mean, I can see the repo | Ok, what are "bin" features? |
| We can rename the channel… | They set the features value to NaN? |
| Um I had issues | Is the fuzzy query bug fixed? |

This study featured a small tagged dataset of 1000 tagged messages per channel. Hence, to avoid overfitting, we limited the number of variables and avoided practices requiring parameter tuning on validation data. A larger set would enable choosing parametric evaluation models as well as providing further validation. Our *Ans-Chat* approach achieved superior results with F-scores of 0.724 and 0.705 on two channels.

# References

Shlomo Berkovsky, Timothy Baldwin, and Ingrid Zukerman. 2008. Aspect-based personalized text summarization. In *Lecture Notes in Computer Science (including subseries Lecture Notes in Artificial Intelligence and Lecture Notes in Bioinformatics)*, volume 5149 LNCS, pages 267–270.

Rose Catherine, Rashmi Gangadharaiah, Karthik Visweswariah, and Dinesh Raghu. 2013. Semi-Supervised Answer Extraction from Discussion Forums. *Proc. IJCNLP 2013*(October):1–9.

Gao Cong, Long Wang, Chin-Yew Lin, Young-In Song, and Yueheng Sun. 2008.

Finding question-answer pairs from online forums. In *Proceedings of the 31st annual international ACM SIGIR conference on Research and development in information retrieval - SIGIR '08*, page 467, New York, New York, USA. ACM Press.

P Deepak and Karthik Visweswariah. 2014. Unsupervised Solution Post Identification from Discussion Forums. *Acl*:155–164.

A. P. Dempster, N. M. Laird, and D. B. Rubin. 1977. Maximum Likelihood from Incomplete Data via the EM Algorithm. *Journal of the Royal Statistical Society. Series B*, 39(1):1–38.

Liangjie Hong and Brian D. Davison. 2009. A classification-based approach to question answering in discussion boards. In *Proceedings of the 32nd international ACM SIGIR conference on Research and development in information retrieval - SIGIR '09*, page 171.

Po Hu, Dong-Hong Ji, Chong Teng, and Yujing Guo. 2012. Context-Enhanced Personalized Social Summarization. In *COLING*, pages 1223–1238.

Adekunle Isiaka Obasa, Naomie Salim, and Atif Khan. 2016. Hybridization of Bag-of-Words and Forum Metadata for Web Forum Question Post Detection. *Indian Journal of Science and Technology*, 8(32), May.

Su Nam Kim, Li Wang, and Timothy Baldwin. 2010. Tagging and Linking Web Forum Posts. *Proceedings of the Fourteenth Conference on Computational Natural Language Learning*(July):192–202.

David D. Lewis and David D. 1992. Feature selection and feature extraction for text categorization. In *Proceedings of the workshop on Speech and Natural Language - HLT '91*, page 212, Morristown, NJ, USA. Association for Computational Linguistics.

J. MacQueen. 1967. eAssistant: Some methods for classification and analysis of multivariate observations. Proc. Fifth Berkeley Symp. on Math. Statist. and Prob. Vol. 1 (Univ. of Calif. Press, 1967), 281--297

Hamid R. Motahari Nezhad, Kalpa Gunaratna, and Juan Cappi. 2017. eAssistant: Cognitive Assistance for Identification and Auto-Triage of Actionable Conversations. *Proceedings of the 26th International Conference on World Wide Web Companion*:89–98.

Jacki O'Neill and David Martin. 2003. Text chat in action. *Proceedings of ACM SIGGROUP Conference on Supporting Group Work*, 33(0):40–49.

Krish Perumal. 2016. *Semi-supervised and Unsupervised Methods for Categorizing Posts in Web Discussion Forums*. Ph.D. thesis.

Krish Perumal and Graeme Hirst. 2016. Semi-supervised and Unsupervised Categorization of Posts in Web Discussion Forums using Part-of-Speech Information and Minimal Features. In *Proceedings of the 7th Workshop on Computational Approaches to Subjectivity, Sentiment and Social Media Analysis*, pages 100–108.

Bernard. W. Silverman. 2018. *Density Estimation for Statistics and Data Analysis*. Routledge, February.

Naama Tepper, Anat Hashavit, Maya Barnea, Inbal Ronen, and Lior Leiba. 2018. Collabot: Personalized Group Chat Summarization. In *Proceedings of the Eleventh {ACM} International Conference on Web Search and Data Mining,*

*{WSDM} 2018, Marina Del Rey, CA, USA, February 5-9, 2018*, pages 771–774.

Naama Tepper, Amir Nisan Cohen, Anat Hashavit, Inbal Ronen, and Lior Leiba. Implicit User Profiling in Multiparticipant Chat. In *UMAP '18 Adjacent proceedings of the 26th Conference on User Modeling, Adaptation and Personalization*.

David C. Uthus and David W. Aha. 2013. Multiparticipant chat analysis: A survey. *Artificial Intelligence*, 199–200:106–121.

David C Uthus and David W Aha. 2011. Plans Toward Automated Chat Summarization. *Computational Linguistics*(Code 5514):1–7.

L.a b Wang, M.a b Lui, S.N.a b Kim, J.c Nivre, and T.a b Baldwin. 2011. Predicting thread discourse structure over technical web forums. *EMNLP 2011 - Conference on Empirical Methods in Natural Language Processing, Proceedings of the Conference*:13–25.

Li Wang, Marco Lui, Su Nam Kim, Joakim Nivre, and Timothy Baldwin. Predicting Thread Discourse Structure over Technical Web Forums. :13–25.

Amy X. Zhang, Bryan Culbertson, and Praveen Paritosh. 2017. Characterizing Online Discussion Using Coarse Discourse Sequences. *Proceedings of the 11th International AAAI Conference on Weblogs and Social Media*(Icwsm):357–366.